\documentclass[journal]{IEEEtran}
\usepackage{graphicx} 
\usepackage{enumitem}
\usepackage{hyperref}
\usepackage{xcolor}
\usepackage{url}
\usepackage[caption=false,font=footnotesize]{subfig}
\usepackage{booktabs}
\usepackage{tabularx}
\usepackage{array}

\title{From State Synchronization to Cognitive Self-Evolution: An Operational Architecture for Cognitive Digital Twins}

\begin{document}

\author{Haoran~Gao, An~Li, Zhen~Li, and~Jun~Cai,~\IEEEmembership{Senior~Member,~IEEE,}\thanks{H. Gao, A. Li, Z. Li and J. Cai are with the Department of Electrical and Computer Engineering, Concordia University, Montreal QC H3G 1M8, Canada (Emails: jun.cai@concordia.ca; haoran.gao@mail.concordia.ca; an.li@mail.concordia.ca; zhen\_li@ieee.org).}}

\maketitle

\begin{abstract}
As Digital Twin (DT) systems evolve beyond state synchronization toward task-oriented and knowledge-driven operation, Cognitive Digital Twins (CDTs) have emerged as an extension that incorporates cognitive capabilities into twin operation. Existing CDT studies often focus on specific enabling techniques, such as learning modules, knowledge graphs, and large language models, while providing limited insight into how cognition can be systematically integrated into DT architectures. To address this issue, this paper proposes a four-layer CDT architecture consisting of the physical layer, digital-twin layer, cognitive layer, and task layer. The proposed architecture establishes a self-evolving closed operational loop spanning these four layers, in which physical states are synchronized into digital representations, cognition constructs task-specific cognitive models through knowledge, memory, and attention, and task-level decisions are generated under practical constraints. Operational feedback further refines cognitive experience and updates relationships and annotations in the digital representation, enabling subsequent task interpretation, initiation, and reasoning to evolve with system operation. Based on this framework, two representative operation modes are characterized: user-request-driven cognition and self-driven cognition. We further discuss key enabling mechanisms and deployment challenges associated with semantic communication, knowledge querying, task orchestration, and closed-loop synchronization. A lightweight simulation study illustrates reliable closed-loop task feasibility under limited semantic information and improved operational efficiency through accumulated task experience. The proposed framework provides a structured foundation for the design and development of future CDT systems.
\end{abstract}


\section{Introduction}
The convergence of physical and digital spaces is fundamentally reshaping the architecture of future intelligent systems~\cite{I1}. In this context, Digital Twin (DT) technology has emerged as a promising paradigm for establishing persistent digital counterparts of physical entities. By maintaining a continuously synchronized virtual representation of a physical object, DTs provide a unified foundation for bridging physical processes and digital intelligence across a wide range of application domains, including smart manufacturing, healthcare, intelligent transportation, and cyber-physical systems~\cite{I2,I4}.

Early developments focused on representing physical entities, while subsequent research has explored their role in supporting operational objectives~\cite{I5}. As a result, task-aware frameworks have emerged, in which synchronized information is incorporated into optimization and learning processes to facilitate task execution~\cite{I6,I7}. This development broadens the function of digital twins from state representation to operational support. For example, fault diagnosis may rely on equipment states, historical failure patterns, and risk-related knowledge, while resource adaptation may depend more strongly on network load, link quality, and latency requirements. Different tasks therefore require different state information, contextual knowledge, and operational objectives. Despite their ability to support operational decision-making, most existing frameworks remain optimization-oriented. Their decisions are typically derived from predefined objectives, rewards, and constraints, implicitly assuming that the significance of system states and operational goals is already understood. In realistic environments, however, predefined objectives are often insufficient to fully characterize task requirements. Determining what information is relevant, which objectives should be prioritized, and how competing factors should be balanced frequently depends on understanding rather than optimization alone. This gap motivates the incorporation of cognitive capabilities that can support interpretation, reasoning, and adaptation within DT systems.

In general, cognition refers to the capability of a system to interpret contextual information, exploit prior knowledge, reason under uncertainty, and adapt behavior based on accumulated experience. Such cognitive mechanisms have gradually appeared in intelligent and human-centered systems, including Human Digital Twin (HDT)-oriented applications, where knowledge integration, reasoning modules, or learning mechanisms are introduced to support contextual interpretation and adaptive decision-making~\cite{I8,I9}. Compared with task-aware DTs that mainly optimize predefined objectives, these systems place greater emphasis on contextual understanding and adaptive reasoning. Building upon this trend, recent studies have further introduced the concept of Cognitive Digital Twins (CDTs), where cognition is incorporated through learning-based adaptation, knowledge-driven reasoning, or large-model-assisted decision support~\cite{I10, I11, I12}. These studies extend DTs beyond state synchronization and passive prediction toward more intelligent decision capabilities. However, cognition is often introduced as an additional decision-support functionality rather than an integral component of DT operation. Consequently, the relationships among physical-state synchronization, cognitive reasoning, task execution, and system evolution remain insufficiently characterized, leaving the role of cognition in the operational workflow of CDT systems largely unclear.

Once cognition is treated as a core operational capability rather than a standalone decision-support tool, several questions emerge. How should cognition interact with state synchronization and task execution? How can task outcomes reshape future cognitive understanding? How should communication, orchestration, and synchronization support the DT closed loop? Existing studies provide limited answers, making it difficult to systematically characterize CDT operation and its unique design requirements.

These questions cannot be answered simply by attaching cognitive modules to existing DT architectures. In conventional DTs, closed-loop operation commonly refers to bidirectional physical--digital interaction, where updated digital states support predefined control, synchronization, or resource adaptation. Such interaction links physical observation with system response, but does not necessarily allow task outcomes to reshape information organization and cognitive reasoning. In contrast, the CDT closed loop extends feedback into cognition itself. Task outcomes refine cognitive experience and subsequent reasoning, while also informing how task-relevant states, contextual conditions, and outcomes are associated and annotated in the digital representation. These feedback-derived representations can then support subsequent cognitive interpretation and autonomous task initiation, allowing cognitive understanding to evolve with accumulated experience. CDT operation therefore forms a continuous process of state synchronization, cognitive interpretation, task execution, and outcome-driven evolution. A unified operational architecture is thus needed to describe how these processes are connected and how cognition participates in system operation over time.

In CDT, physical-state generation, digital-state representation, cognitive interpretation, and task-oriented decision making correspond to four distinct operational functions in the CDT lifecycle. Explicitly separating these functions helps clarify their responsibilities and interactions within the cognitive feedback process. Thus, a four-layer architecture is adopted to organize these functions within a unified operational framework. Within this architecture, two representative operation modes are considered: explicit task triggering from the physical layer and autonomous task generation by the cognitive layer upon detecting abnormal twin states. Through this task-oriented cognitive feedback architecture, CDT is formulated as a closed-loop operational system that supports continuous adaptation and self-evolution during operation. The main contributions of this paper are summarized as follows.
\begin{itemize}
    \item  This work provides an operational definition of CDT by distinguishing it from general cognition-enhanced DTs, where physical synchronization, cognitive interpretation, task execution, and feedback-driven evolution are integrated into a unified closed-loop architecture.
    \item  This work identifies two fundamental pathways through which cognition enters CDT operation: user-request-driven cognition and self-driven cognition, revealing how explicit task requests and evolving synchronized states respectively initiate task-oriented reasoning.
    \item  The deployment characteristics and challenges of CDT architectures under end-edge-cloud environments are further examined. Representative case studies are conducted to validate the feasibility and search-efficiency advantages of the proposed CDT framework.
\end{itemize}

The remainder of this paper is organized as follows. Section II reviews the related work. Section III introduces the system model together with its major components, the associated design challenges, and the key techniques that enable the proposed framework. Section IV reports the simulation results. Finally, Section V concludes this research.

\begin{table*}[!t]
\centering
\caption{Comparison of representative digital twin paradigms.}
\renewcommand{\arraystretch}{2}
\setlength{\tabcolsep}{4pt}
\begin{tabularx}{\textwidth}{
    >{\raggedright\arraybackslash}m{2.8cm}
    >{\raggedright\arraybackslash}X
    >{\raggedright\arraybackslash}X
    >{\raggedright\arraybackslash}X
    >{\raggedright\arraybackslash}X
}
\toprule
\textbf{Dimension}
&
\textbf{DT}
&
\textbf{HDT}
&
\textbf{Existing CDT}
&
\textbf{Proposed CDT}
\\
\midrule

\textbf{System structure}
&
Digital representation + Prediction function
&
DT + External static knowledge base + Task-output modules
&
End-to-end black box, or explicit only in cognitive module
&
Four-layer: physical, digital-twin, cognitive, and task layers
\\
\midrule

\textbf{Cognitive process}
&
Not modeled
&
External knowledge-base retrieval only
&
Black-box, or task-related knowledge retrieval
&
Combines task requirements, synchronized states, relevant knowledge, and feedback from previous task execution
\\
\midrule

\textbf{Task generation}
&
Not modeled
&
Triggered by external healthcare or service demands
&
User-request-driven
&
User-request-driven or self-driven
\\
\midrule

\textbf{Use of outcomes}
&
State estimation and prediction
&
Task-oriented assessment or intervention recommendations
&
Task solutions with reward- or model-level feedback
&
Task solutions with refinement of memory, knowledge, attention, and subsequent reasoning
\\

\bottomrule
\end{tabularx}
\end{table*}

\section{Related Work}
Existing studies on digital representations of physical systems can be broadly categorized into three directions: traditional digital twins (DTs), human digital twins (HDTs), and cognitive digital twins (CDTs). This section reviews these representative paradigms and clarifies the gap between existing DT formulations and the operational CDT architecture considered in this paper.

\subsection{Traditional Digital Twin (DT)}
DTs were originally introduced as high-fidelity virtual representations of physical objects, maintained through continuous data synchronization to support monitoring, diagnosis, and prediction. In industrial systems, DTs have been widely used to model machines, production lines, and infrastructure, enabling predictive maintenance and simulation-based analysis of system dynamics~\cite{I2}. In wireless networks, the DT concept has further evolved into digital twin networks (DTNs), where virtual representations of network entities such as base stations, satellites, edge devices, and communication links are constructed to support network monitoring, resource management, and performance prediction~\cite{R1}.

Despite these advances, existing TDT frameworks still mainly focus on physical-state representation, synchronization, and evolution analysis. Although such digital representations provide an important foundation for system analysis, they offer limited support for task-oriented operation, constraint-aware decision-making, and knowledge-driven closed-loop adaptation. As a result, TDTs provide the entity-level digital representation required by complex systems, but do not by themselves constitute an operational cognitive framework.

\subsection{Human Digital Twin (HDT)}
Human-centered applications further demonstrate that digital twin systems often need to support not only state representation, but also task-oriented interpretation assisted by domain knowledge. In medical and healthcare scenarios, twins are expected to identify abnormal physiological patterns, associate them with clinical knowledge, and support diagnostic or intervention decisions. In this sense, HDTs can be regarded as an intermediate form between state-centric DTs and CDTs, since they extend DTs from physical-state synchronization toward human-centered task analysis supported by domain knowledge. HDTs integrate multimodal physiological signals, clinical records, behavioral information, and contextual factors to construct virtual representations of individuals for risk assessment, disease progression prediction, and personalized treatment. Compared with infrastructure-oriented DTs, HDTs are more closely associated with human-centered tasks, where numerical prediction alone is often insufficient. For example, predicted physiological changes typically need to be interpreted together with medical rules, patient history, and contextual conditions before meaningful intervention can be generated. Consequently, many HDT frameworks incorporate medical knowledge bases, ontologies, or expert rules to improve the interpretability of twin states and align analytical results with clinical practice~\cite{I8,I9,R3}.

However, in most HDT frameworks, domain knowledge is still introduced as an external support module rather than as an endogenous cognitive layer within the twin. Although such systems move beyond pure state replication and reveal the need for task-aware interpretation and cognitive functionalities, they still lack a closed-loop cognitive architecture in which knowledge, memory, task generation, and feedback-driven evolution are continuously integrated into the twin operation process.

\subsection{Cognitive Digital Twin (CDT)}
The term CDT has been used in different studies to describe DT systems enhanced with cognitive capabilities. However, the meaning of ``cognition'' remains ambiguous in much of the existing literature~\cite{R4}. In some studies, cognition mainly refers to adaptive control or online policy optimization enabled by reinforcement learning or deep learning modules~\cite{I11}. In others, it denotes knowledge-driven interpretation supported by expert systems, rule-based reasoning, or knowledge graphs~\cite{I12}. More recently, large language models and foundation models have also been incorporated into DT systems to support natural-language interaction, report generation, and decision assistance~\cite{I10}. Although these studies introduce different forms of intelligence into DT systems, cognition is often characterized through the adopted technique rather than through a unified operational mechanism.

A common limitation of existing CDT formulations is that cognition is often treated as an auxiliary functionality rather than an endogenous part of the twin operation process. Consequently, the interaction among cognition, physical--digital synchronization, task execution, and operational feedback remains loosely defined, and CDT operation is rarely modeled as a continuously evolving closed-loop process.

\begin{figure*}[!t]
  \centering
  \subfloat[CDT operation under user-request-driven cognition.]{
    \includegraphics[width=0.48\linewidth]{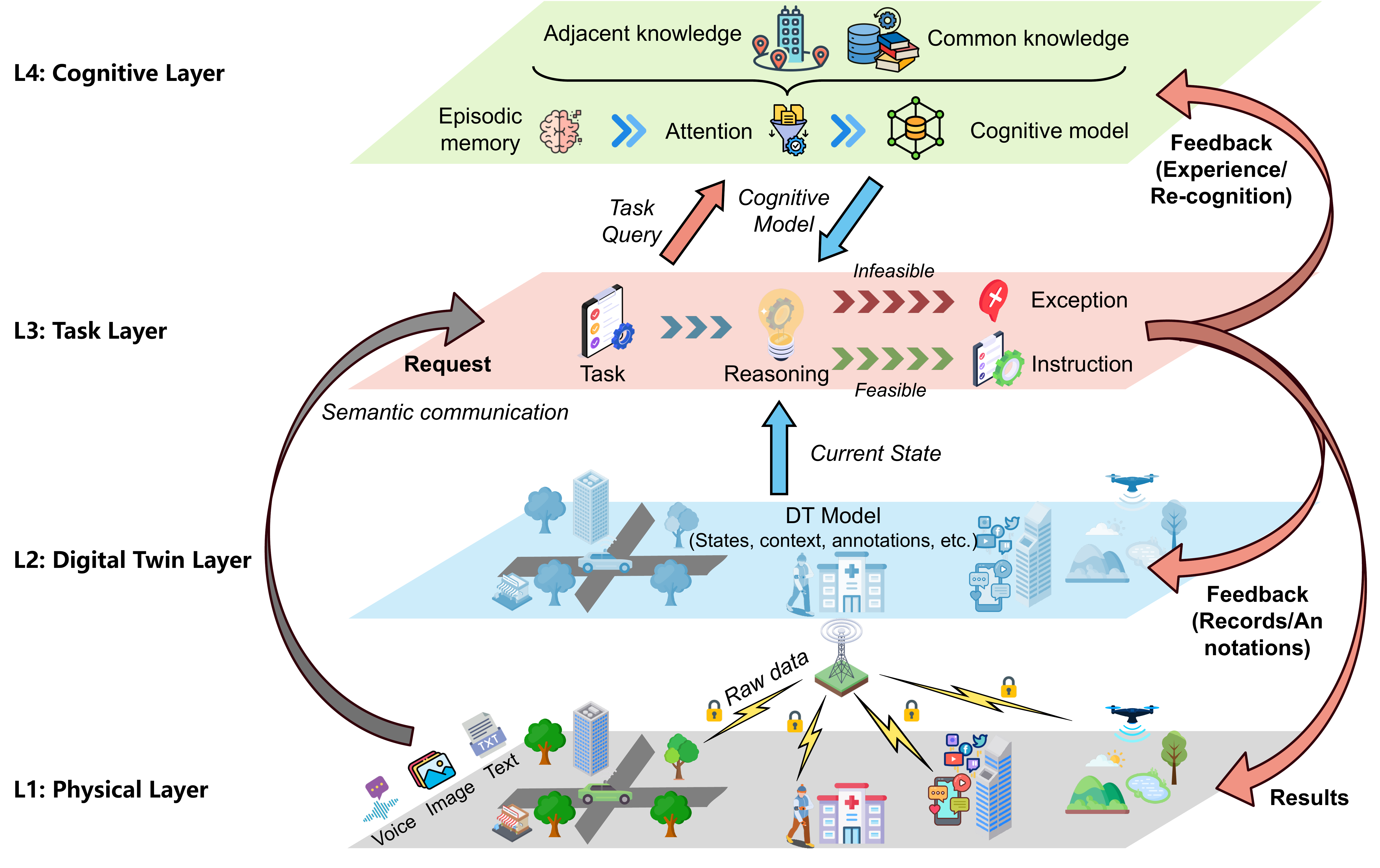}
  }\hfill
  \subfloat[CDT operation under self-driven cognition.]{
    \includegraphics[width=0.48\linewidth]{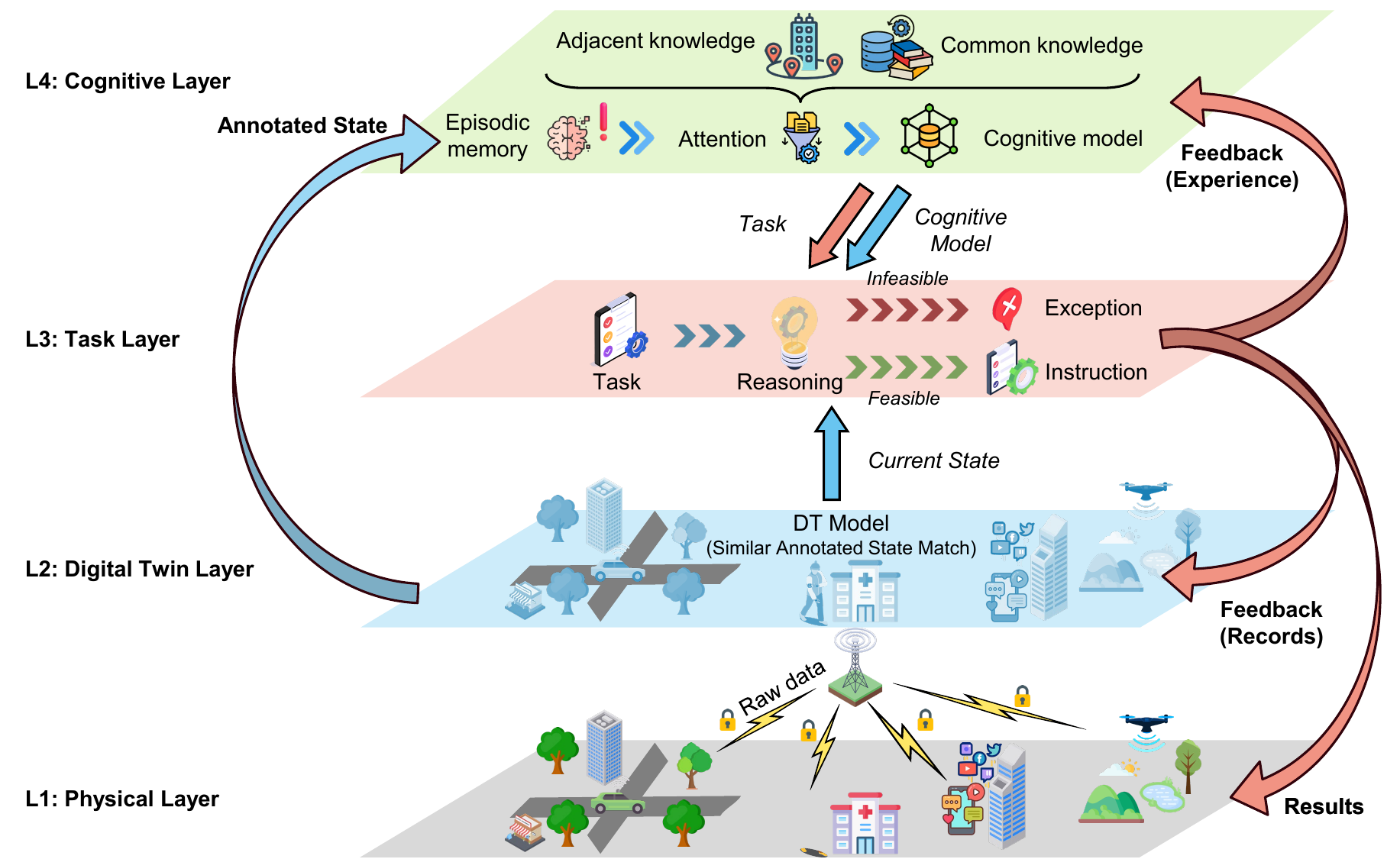}
  }
  \caption{Four-layer closed-loop cognitive digital twins architecture.}
\end{figure*}

From this perspective, self-evolution should be regarded as a fundamental property of CDT rather than an optional extension. While conventional DT systems continuously update their digital states in response to changes in their physical counterparts, CDT systems should further refine their cognitive understanding based on task outcomes, feedback signals, and accumulated operational experience. Without this capability, the system remains a DT assisted by external cognitive tools rather than a genuinely self-evolving cognitive twin. CDT should therefore be formulated as a closed-loop operational architecture in which physical states are synchronized into the digital space, knowledge and memory support the interpretation of states and tasks, task-oriented actions are generated under practical constraints, and execution feedback continuously refines subsequent cognitive understanding. Table I summarizes the distinctions among DT paradigms and highlights the characteristics of the proposed CDT architecture.

\section{An Outline of CDT Proposed Architecture}
In this section, we propose a four-layer operational architecture for CDT, as illustrated in Fig. 1. The architecture characterizes how physical-state synchronization, cognitive interpretation, task execution, and outcome feedback interact to form a closed-loop operational system. We further illustrate its operation through two representative task-initiation modes and summarize the key requirements, design challenges, and potential enabling technologies for CDT realization.

\subsection{Four-Layer System Architecture}
We consider a task-oriented CDT associated with a single physical entity in an end-edge-cloud environment. Rather than serving as a passive data source, the physical entity continuously generates evolving states and task-driven operational demands that shape CDT behavior. In practical scenarios, the CDT may support abnormal-state assessment, intervention planning, resource adaptation, fault diagnosis, or risk warning under dynamic environmental and communication constraints. The physical layer (\textit{P-layer}) comprises the physical entity and its surrounding environment. It provides sensor observations, contextual information, operational logs, and task-related requests for CDT synchronization, while receiving control actions, alert signals, or decision recommendations generated by the CDT and applying them to the physical world. The P-layer therefore serves as the interface for continuous interaction between the CDT and the physical entity.

On top of the physical layer, the digital twin layer (\textit{DT-layer}) maintains a synchronized digital representation of the physical entity based on the physical states generated from the P-layer. Beyond state synchronization, the DT-layer integrates current states, contextual information, and operational records into a structured representation that reflects the operating condition of the physical entity. This representation preserves the state information and provides the digital basis for subsequent cognitive interpretation. The DT-layer itself does not determine which information is relevant to a particular task or assign semantic meaning to the observed states. Nevertheless, the DT-layer can record the outputs of cognitive and task-execution processes, including identified relationships among physical states, contextual conditions, and task outcomes, together with annotations. These annotations retain the results of previous cognitive interpretation, enabling similar future states to trigger further cognitive processing. In this way, cognitive interpretation can be reflected in subsequent digital representations without transferring the interpretation function itself to the DT-layer.

The cognitive layer (\textit{C-layer}) hosts the cognitive mechanisms of the CDT. It combines specified task demands with needs internally identified from synchronized states, such as emerging risks, objective deviations, or other task-relevant conditions. Based on the structured digital representation provided by the DT-layer, the C-layer constructs a task-specific cognitive model by jointly considering the task or operational need, synchronized states, relevant knowledge, and experience derived from task outcomes. Memory retains task-relevant experience learned from prior cognitive and task-execution processes; attention selects the contextual conditions, experience, and knowledge elements relevant to the current situation; and knowledge provides the concepts, relationships, rules, and constraints required for interpretation, drawing on both locally maintained and adjacent shared knowledge when available. Through their interaction, these mechanisms form a task-specific cognitive model that captures the semantic interpretation, relevant relationships, and applicable cognitive constraints associated with the task. For example, in a health-oriented CDT, an eye-related task may require the joint interpretation of the observed eye condition, associated symptoms, environmental context, previous experience, and relevant medical knowledge. The resulting cognitive model is then provided to the task layer for task-oriented reasoning and decision-making. Task outcomes subsequently refine the task-relevant experience retained in memory, the associations represented in knowledge, and the information-selection patterns of attention. These updates also provide the relationships and annotations recorded by the DT-layer, allowing subsequent cognition to build on previous operation. Thus, cognition functions as an endogenous and self-evolving process within the CDT rather than as an external intelligent module attached to the twin.

The task layer (\textit{T-layer}) is responsible for task-oriented reasoning, planning, and decision making. Given the task objective, the structured digital representation provided by the DT-layer, and the task-specific cognitive model generated by the C-layer, the T-layer translates the objective into operational requirements and constructs a set of candidate actions or plans. The cognitive model provides task-relevant interpretations, relationships, priorities, and constraints that guide candidate generation. Each candidate must also be executable under the physical and operational conditions represented by the DT-layer. The T-layer evaluates these candidates and selects the one that best satisfies the task objective. If no candidate can satisfy the cognitive requirements under the available state information and execution conditions, the task is considered infeasible, and an exception is generated to indicate the missing information, conflicting requirements, or unavailable resources that prevent task completion. This exception is returned to the C-layer as feedback, prompting it to revise the task-specific cognitive model by reconsidering the relevant information, relationships, priorities, and constraints.

Overall, the four layers form a self-evolving closed-loop system. The P-layer provides physical states and external task requests when applicable. The DT-layer maintains a structured digital representation and records feedback-derived relationships and annotations that support task-relevant pattern identification and the initiation of self-driven cognition. The C-layer combines external requests or internally identified needs with memory, attention, and knowledge to construct a task-specific cognitive model. The T-layer then uses this model and the digital representation to generate and evaluate candidate actions, returning either an executable instruction or an exception. Task outcomes and exceptions refine cognitive experience in the C-layer and update the records maintained by the DT-layer. Through this loop, cognition and execution continuously shape subsequent interpretation, task initiation, and decision making, enabling self-evolving CDT operation.

\begin{figure}[!t]
  \centering
  \includegraphics[width=\linewidth]{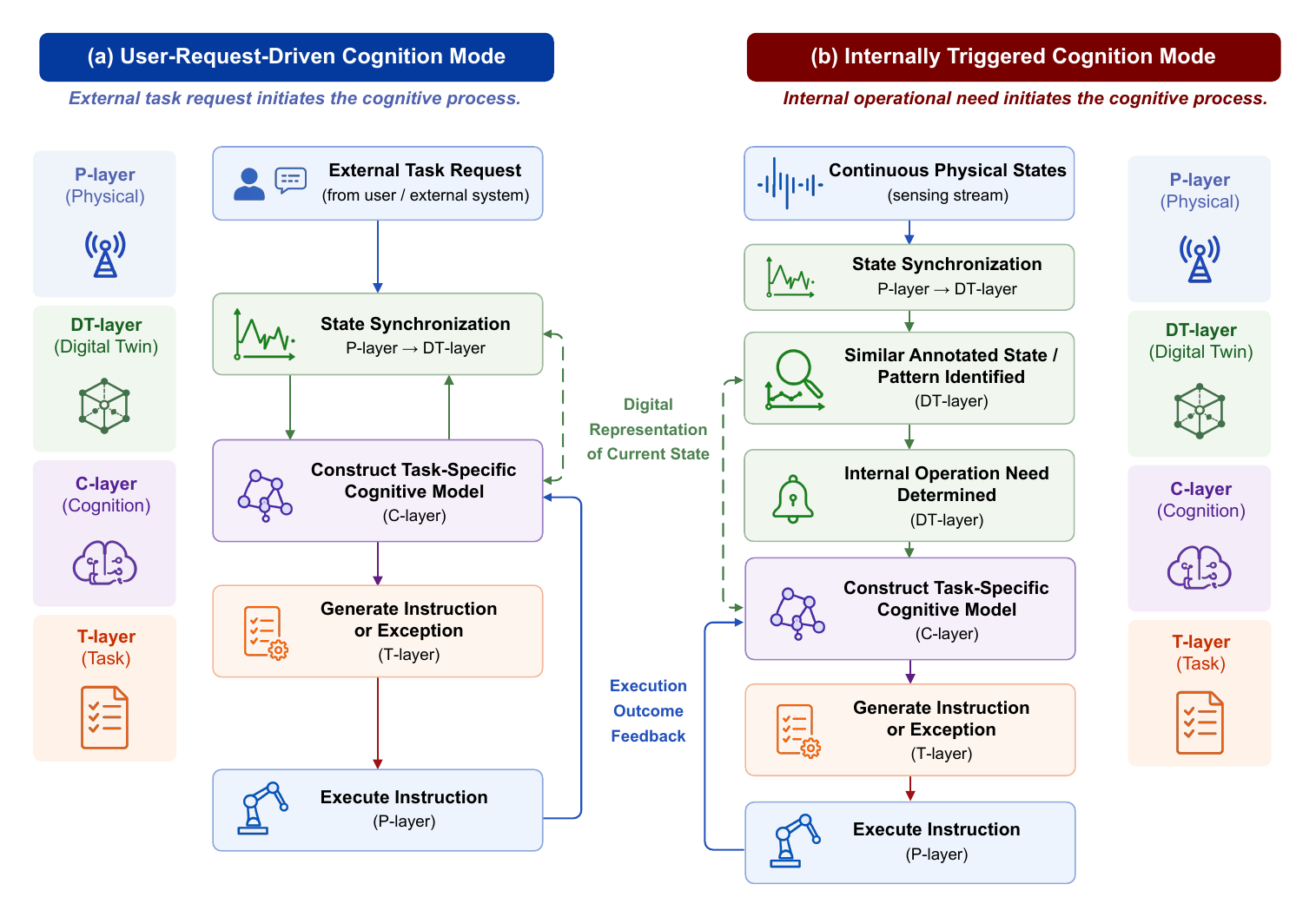}
  \caption{Two operation modes of the proposed CDT architecture.}
\end{figure}

\subsection{Operation Modes of CDT}
While the proposed four-layer architecture defines the functional organization of CDT, an important question concerns how the cognitive closed loop is activated. Unlike conventional DT systems, where tasks are typically predefined and externally assigned, CDT systems allow task initiation to arise from different sources. Tasks may be triggered either by explicit requests received through the physical layer or by cognitive interpretation of evolving system states and task-relevant patterns. Accordingly, we characterize two CDT operation modes: user-request-driven cognition and self-driven cognition.

\textit{Case 1:} External tasks are explicitly initiated through the P-layer via semantic communication, as illustrated in Fig.~1(a). The P-layer constructs a task request from inputs such as text, audio, or images for purposes including physical assessment, service adjustment, or intervention recommendation. Instead of transmitting all raw observations, the request is encoded into a semantic representation containing the task intent and key contextual information. Upon receiving the request, the C-layer combines the task intent with relevant knowledge and prior experience to construct a task-specific cognitive model. Meanwhile, the DT-layer provides the digital representation of the physical entity, including the state and operational conditions required for task execution. The T-layer then combines the cognitive model with this representation to generate and evaluate candidate actions under the task objective and practical constraints. If a feasible action exists, the corresponding instruction or recommendation is returned to the P-layer. Otherwise, an exception is returned to the C-layer, prompting revision of the cognitive model by reconsidering the selected information, relevant relationships, or applicable constraints. Task outcomes and exceptions further refine the cognitive experience retained by the C-layer, including which information was relevant, which knowledge and relationships supported the task, which constraints shaped the decision, and why the selected action succeeded or failed. They also update the records and annotations maintained by the DT-layer. Thus, this mode realizes a task-driven cognitive closed loop rather than a direct input-output inference process.

\textit{Case 2:} Tasks are internally initiated from evolving system states, as illustrated in Fig.~1(b). In this mode, the physical entity does not issue an explicit task request but continuously provides physical states through the P-layer, which are synchronized and recorded by the DT-layer. When a newly observed state matches or resembles a previously annotated state pattern, the DT-layer forwards the corresponding state information and annotations to the C-layer for further interpretation. The C-layer combines this information with relevant knowledge and prior experience to determine whether an operational need exists and, if so, constructs the corresponding task-specific cognitive model. The subsequent decision-making, outcome generation, and feedback processes follow those described in Case 1. Thus, this mode enables previous cognitive and task-execution results to support autonomous task initiation when similar state conditions emerge.

These two modes reflect the request-driven and autonomous characteristics of the proposed CDT. In the user-request-driven mode, cognition is activated by an explicit semantic task request, while the DT-layer provides the current operational representation required for task execution. In the self-driven mode, cognition is initiated when evolving states correspond to previously annotated patterns, allowing prior cognitive and task-execution outcomes to support the identification of new operational needs. Although their initiation mechanisms differ, both modes follow the same subsequent process: the C-layer constructs a task-specific cognitive model, and the T-layer combines this model with the current digital representation to generate an executable action or exception. The distinct initiation mechanisms and shared subsequent processing of the two modes are illustrated in Fig. 2. Together, these modes demonstrate that the proposed CDT is not a static DT assisted by an external intelligence module, but an integrated cognitive system in which external requests, state evolution, and operational feedback jointly drive task execution and self-evolution.

\subsection{Design requirements and challenges}
By introducing explicit cognition and task-level closed loops, CDT extends beyond the state-estimation paradigm of conventional DTs and introduces new design requirements and challenges, discussed below.

\begin{itemize}
    \item[1)] \textbf{Task-aware digital-state organization}: A practical challenge is determining which parts of the digital representation should support a cognitive process and how their relevance should be reflected in the DT-layer. In real-world scenarios, a digital twin may contain numerous states, contextual attributes, and historical records that are difficult to synchronize and use simultaneously. Different tasks, however, often depend on different information subsets, whose relevance may change as environmental conditions evolve. Excessive information may introduce unnecessary communication, computation, and reasoning overhead, whereas insufficient information may lead to incomplete cognitive interpretation and unreliable decisions. Thus, the DT-layer requires mechanisms to organize and activate states and contextual information according to task-relevance relationships and annotations.
    \item[2)] \textbf{Semantic relevance requirement}: CDT introduces task-oriented semantic communication from the P-layer to the C-layer to convey user requests and information relevant to decision making. In wireless environments with limited bandwidth and stringent latency requirements, the P-layer must selectively transmit information that can influence cognitive-model construction in the C-layer and subsequent task planning in the T-layer. However, semantic relevance is task dependent, making it difficult to define quantifiable measures of semantic importance for different CDT tasks~\cite{R3}. Meanwhile, high-level semantic representations must be mapped into transmittable bit streams and, after decoding, preserve structures and meanings useful for cognitive-model construction and task-level decision making. Realizing this round-trip mapping between semantics and bits under wireless constraints, without excessive distortion or latency, represents another distinctive challenge for CDT compared with DTs relying solely on bit-level communication.
    \item[3)] \textbf{Resource allocation and networking}: Task execution introduces constraints related to cognitive quality (QoC). Unlike DT systems that mainly optimize quality of service (QoS) or quality of networking (QoN), task feasibility in CDT depends on whether the structured digital representation and task-specific cognitive model can be delivered and jointly used by the T-layer within the required time and resource constraints. In multi-task scenarios, tasks differ in priority, cognitive requirements, and feasible domains, while high-priority tasks demand stronger communication and computation support for state synchronization, cognitive-model construction, and task evaluation. Resource allocation therefore becomes a task-aware coordination problem that jointly balances QoS, QoN, and QoC, creating new resource and network design challenges for CDT.
    \item[4)] \textbf{Privacy risks}: When CDTs share or request knowledge over wireless links, privacy leakage and eavesdropping become unavoidable risks. Adjacent knowledge is an integral part of the C-layer and is often closely tied to specific behaviors and decision strategies. Once intercepted or correlated, such knowledge may reveal individual behavior patterns or system-level policies. A key CDT challenge is therefore to limit leakage during adjacent knowledge exchange without compromising task utility.
    \item[5)] \textbf{Efficiency of loops}: CDT forms a cognitive closed loop across the P-, DT-, C-, and T-layers, requiring the wireless network to satisfy end-to-end latency and jitter constraints across multiple interactions rather than isolated transmissions. The network must also maintain consistency among the physical state, digital representation, and task-specific cognitive model during decision making. Defining latency budgets, coordinating cross-layer synchronization and cognitive-model updates, and preserving closed-loop stability and reliability without excessive communication and computation overhead therefore constitute a major CDT challenge.
    \item[6)] \textbf{Incentive mechanisms for adjacent knowledge}: CDT relies on cross-CDT knowledge interaction to support adjacent cognition and task-oriented reasoning. However, in non-trust environments, participants may lack incentives to contribute high-value knowledge and cognitive context. Designing measurable contribution metrics and incentive mechanisms for reliable, high-quality knowledge sharing therefore constitutes a distinctive challenge in CDT protocol and mechanism design.
\end{itemize}

\subsection{Key Techniques}
These challenges show that CDT is not a DT augmented with intelligent modules. Cognition introduces dependencies among state synchronization, knowledge reasoning, task execution, and feedback-driven evolution that exceed the scope of conventional DT solutions. They therefore motivate techniques for task-oriented cognition and self-evolving closed-loop operation. The following discussion outlines representative techniques aligned with the proposed CDT architecture.
\begin{itemize}
    \item[1)] \textit{Task-aware consistency and coordination}: Task-aware digital-state organization should ensure that states and contextual information identified as task relevant through cognitive feedback are properly synchronized, associated, and maintained for CDT operation. Four directions can support this goal. Relevance propagation uses task-relevance relationships and annotations to identify states associated with the current task and extend relevance to related contextual information, avoiding isolated state selection. Prediction and age-of-information (AoI) modeling assess whether infrequently used states may become relevant to future tasks and whether their synchronization should be maintained. Graph-based modeling captures dependencies among states, contexts, and historical records to support coherent organization. Adaptive mechanisms further adjust state organization as task requirements and environmental conditions evolve, allowing the CDT to refine its representations across operational scenarios.
    \item[2)] \textit{Representativeness and structured mapping}: “Key information” in CDT refers to semantic fragments that capture task intent and contextual information under bandwidth and latency constraints while influencing attention allocation and task planning. Semantic criticality can therefore be characterized from two complementary perspectives. Statistical representativeness uses indicators such as deviation, novelty, or uncertainty reduction to preserve information reflecting state evolution and other task-relevant conditions. Task representativeness captures the potential impact of a semantic item on feasible-set construction and policy selection~\cite{R3}. Semantic transmission should also preserve structured meaning when mapping key information into continuous or discrete representations for semantic-to-bit transmission over wireless channels.
    \item[3)] \textit{Cross-dimensional optimization technique}: In CDT, resource allocation must satisfy coupled QoS, QoN, and QoC constraints. Task priority, end-to-end latency and reliability, computational resources, and cognitive effectiveness should therefore be incorporated into a unified framework that explicitly models cross-dimensional tradeoffs or imposes hard guarantees on critical constraints. This enables coordinated optimization of resource allocation and cognitive objectives. Methodologically, the problem can be formulated as a dynamic multi-objective, multi-constraint optimization problem and addressed through online optimization, reinforcement learning, or multi-agent methods.
    \item[4)] \textit{Confidentiality-preserving knowledge exchange}: To address privacy requirements, CDT requires a semantic-confidentiality-preserving mechanism for knowledge exchange. The core idea is to transform exposed wireless transmission into a controlled process that constrains semantic disclosure and cross-context linkability. Shared knowledge should be recoverable only within authorized cognitive and task contexts and remain difficult to exploit for behavioral or policy inference. This mechanism jointly protects semantic confidentiality and limits information visibility, reducing knowledge leakage and the exposure of behavioral patterns and system strategies under passive eavesdropping.
    \item[5)] \textit{Cross-layer closed-loop co-design}: The cognitive closed loop in CDT requires joint coordination across network transmission, state synchronization, cognitive-model construction, and task decision making. Specifically, the forward path couples semantic request delivery and digital-state synchronization with cognitive-model construction and candidate evaluation, while the feedback path couples task outcomes and exceptions with cognitive refinement and DT annotations. Cross-layer co-design should therefore allocate stage-wise latency and reliability budgets according to task deadlines, while ensuring that the digital representation and cognitive model used by the T-layer remain consistent in time and task context. When predicted delay or representation inconsistency threatens task feasibility, synchronization frequency, knowledge querying, network scheduling, and computation placement should be jointly adjusted. In this way, individual layers serve the system-level objectives of closed-loop stability and decision reliability rather than optimizing isolated links or processing stages.
    \item[6)] \textit{Contract-based incentive mechanisms}: In trust-limited and resource-constrained environments, sustaining CDT participation in high-value knowledge sharing is important. Contract theory provides a systematic framework for aligning participants’ private utilities with system-level cognitive objectives. By balancing contribution costs and rewards through contractual protocols, the system can incentivize truthful, high-quality knowledge sharing without compromising overall efficiency~\cite{S1}.
\end{itemize}

\section{A Case Study of CDT}
In this section, a case study is used to illustrate the feasibility and self-evolving characteristics of the proposed CDT architecture. To this end, we focus on two fundamental properties that are central to CDT operation. The first is closed-loop task feasibility, which evaluates whether the CDT can successfully construct feasible task solutions under constrained semantic information. The second is experience-driven evolution, which evaluates whether accumulated operational experience can improve reasoning efficiency and reduce task-processing overhead. Together, these two properties reflect the capability of the proposed architecture to support reliable cognitive operation and continuous self-improvement.

Based on these objectives, a lightweight simulation study is conducted to investigate the runtime behavior of the proposed CDT architecture under closed-loop task execution. In this setup, the end side corresponds to the P-layer that generates sensing data and semantic task requests, the edge side hosts the DT-layer and T-layer for state synchronization and task planning, and the cloud side hosts the C-layer, where a KG-based knowledge repository and shared knowledge are maintained to support task-specific cognitive-model construction. The semantic ratio reflects limited end–edge information transmission, while the search overhead reflects the cost of edge–cloud knowledge querying and candidate-solution searching. Rather than constructing a complex domain-specific twin, this study focuses on evaluating the overall operation process of the proposed architecture under a unified task workflow. For comparison, we consider two baselines:
\begin{enumerate}[label=\arabic*)]
    \item \textbf{CDT without Adjacent Knowledge (CDT-NA)}: This configuration is similar to the proposed CDT operating in the user-request–triggered mode, except that the cognitive layer relies solely on locally accumulated knowledge and does not access adjacent shared knowledge.
    \item \textbf{DT with Static Knowledge-base (DT-S)}: This baseline uses synchronized DT states and a predefined static knowledge base for task handling and solution selection. Unlike the proposed CDT, its knowledge structure and task-processing strategy are not updated based on task outcomes or accumulated experience.
\end{enumerate}

\begin{figure}[!t]
  \centering
  \includegraphics[width=\linewidth]{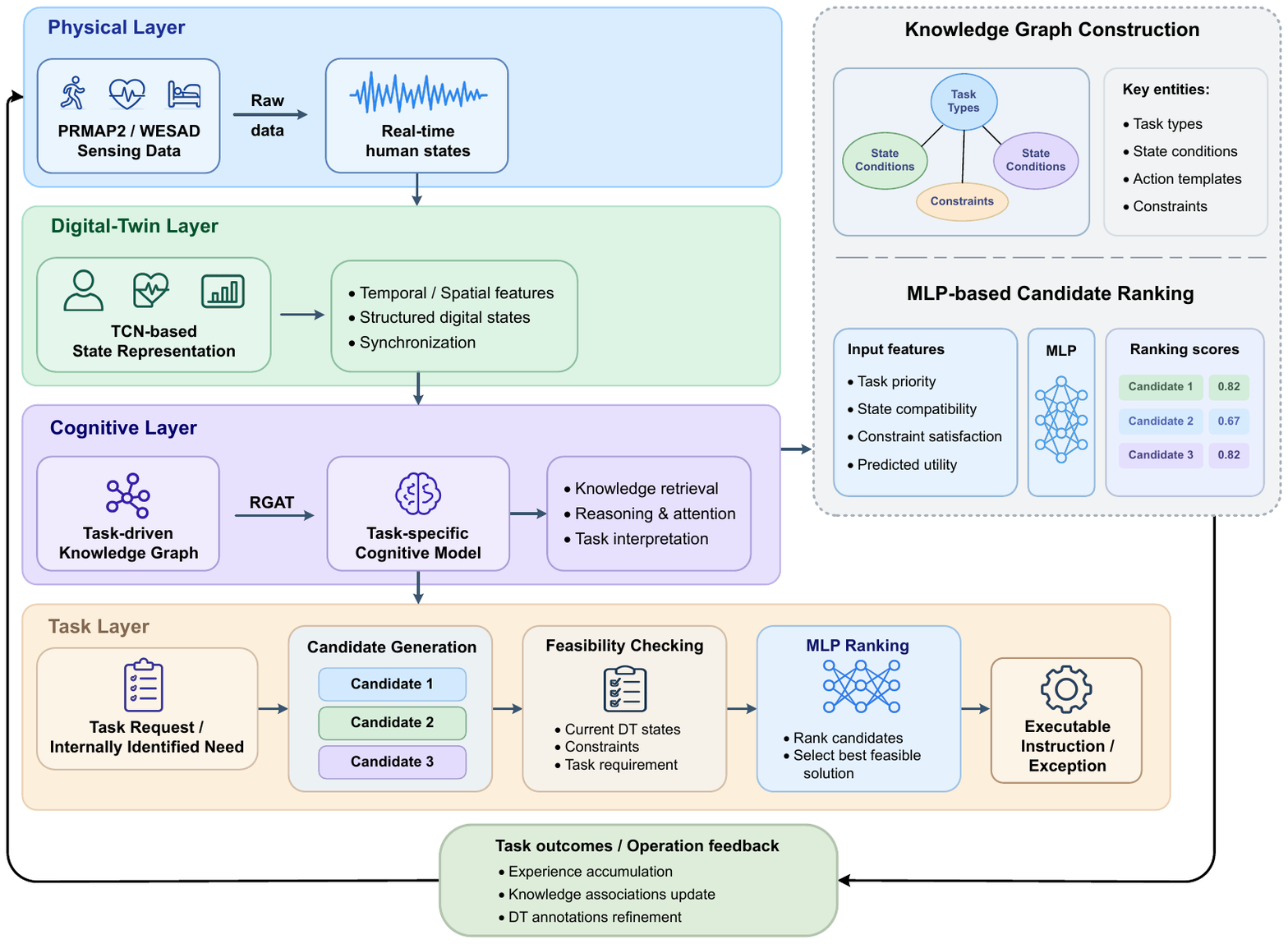}
  \caption{Case-study workflow for cognitive task processing.}
\end{figure}

\subsection{Simulation Settings}
To support the experimental evaluation, we combine two wearable and physiological sensing datasets from the UCI Machine Learning Repository, namely \textit{PAMAP2} and \textit{WESAD}, to construct a representative multivariate state space for DT state synchronization and task-driven closed-loop simulation. Specifically, the PAMAP2 dataset consists of motion and physiological signals collected from multiple wearable sensors, covering diverse activity types and their temporal evolution, while the WESAD dataset provides multimodal physiological signals characterizing individuals’ dynamic responses under different stress and affective conditions.

Meanwhile, the DT-layer adopts a Temporal Convolutional Network (TCN) with causal and dilated convolutions for low-latency temporal state mapping while preserving causality and short-horizon dynamics. In the C-layer, a task-driven knowledge graph (KG) encodes task types, action templates, relevant relationships, and cognitive constraints. Given an external task request or an internally identified operational need, a relational graph attention network (RGAT) retrieves task-relevant knowledge from the KG and constructs the task-specific cognitive model. The T-layer combines this model with the current digital representation maintained by the DT-layer to generate feasible candidate actions, which are ranked by a lightweight multilayer perceptron (MLP) to select the final executable instruction. To evaluate scalability, we vary the number of task requests in the task pool and measure the average search overhead required to identify the first feasible solution. Each task combines a task type, state-dependent constraints, and a candidate action template, with different priorities, deadlines, or state conditions. Increasing the number of tasks therefore enlarges both the search space and the diversity of task conditions. The measured search overhead reflects the runtime cost of traversing candidate actions before obtaining a feasible instruction. The case-study workflow is illustrated in Fig. 3, including task-relevant knowledge retrieval, candidate generation and feasibility evaluation, and MLP-based ranking for instruction selection.

\begin{figure}[!t]
    \centering
    \includegraphics[width=0.48\textwidth]{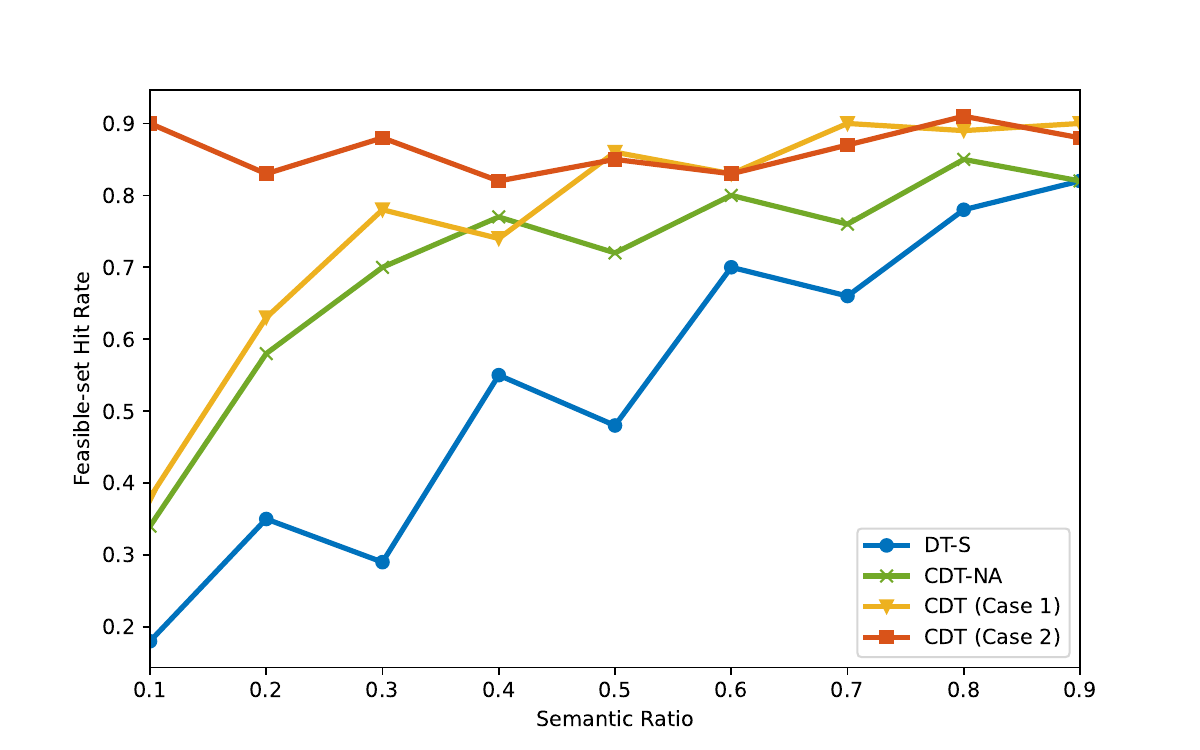} 
    \caption{Feasible-set hit rate under different semantic Ratios.}
\end{figure}

\begin{figure}[!t]
    \centering
    \includegraphics[width=0.48\textwidth]{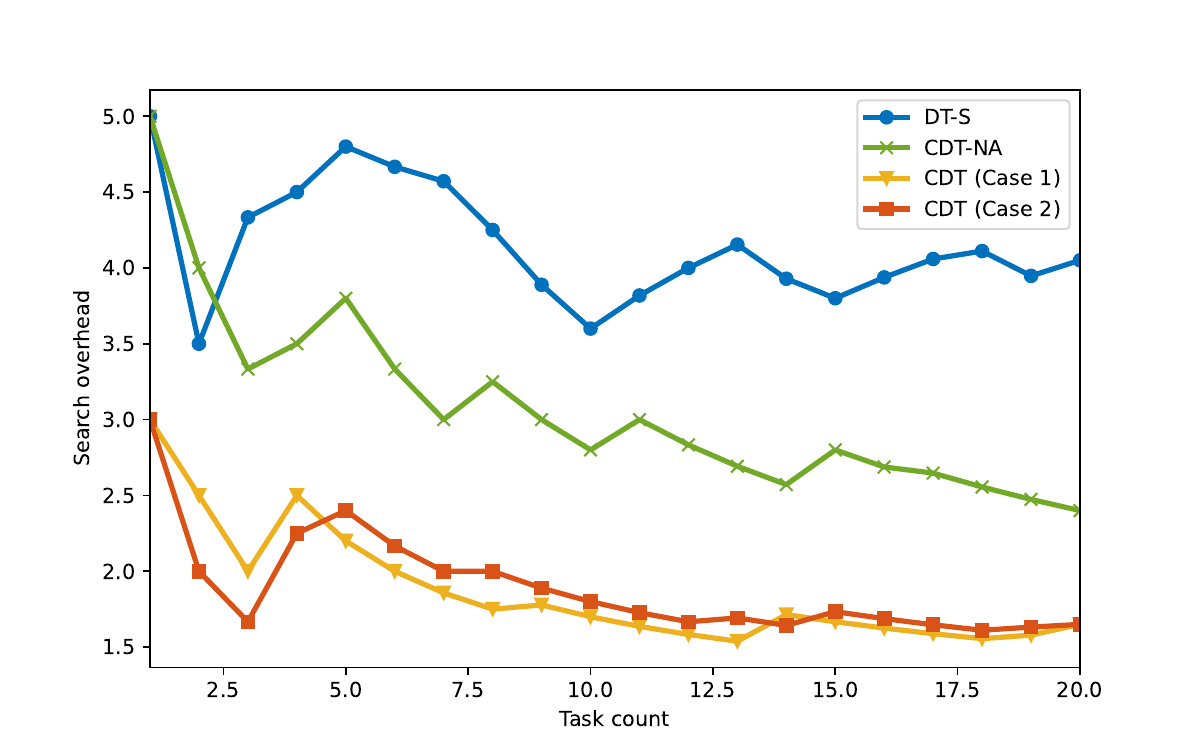} 
    \caption{Search overhead under different number of tasks.}
\end{figure}

\subsection{Evaluation Results}
Both cases of the proposed CDT are considered. Fig. 4 evaluates the closed-loop task feasibility of the proposed CDT architecture under different semantic ratios. To this end, we adopt the feasible-region hit rate as the evaluation metric, where a successful hit indicates that the CDT can construct at least one feasible solution under the available semantic information and operational constraints. The semantic ratio denotes the proportion of task-relevant semantic information retained during transmission from the P-layer for subsequent cognitive and task processing. From a wireless systems viewpoint, the semantic ratio reflects the amount of effective semantic information retained under a given bandwidth and latency budget; thus, a lower semantic ratio indicates that more task-relevant information may be discarded during semantic encoding, and vice versa~\cite{R3}. The feasible-region hit rate reflects the system’s ability to construct a feasible solution space at the initial stage of task processing, indicating whether the available semantic information is sufficient to guide the system toward at least one constraint-satisfying solution under partial observability. The results show that CDT (Case 2) maintains a high and stable hit rate, as task initiation in the self-driven mode relies on synchronized state patterns and feedback-derived annotations rather than on an externally transmitted semantic request. In contrast, CDT (Case 1) is more affected by information loss at low semantic ratios, but its hit rate improves significantly as richer semantic information enables more accurate task-specific cognitive-model construction and candidate-solution generation, eventually approaching the performance of CDT (Case 2). DT-S exhibits the lowest hit rate across all semantic ratio settings. Overall, these results demonstrate that the proposed CDT can maintain reliable closed-loop task feasibility under constrained semantic transmission, while showing the robustness of self-driven operation to variations in transmitted semantic information.

Fig. 5 evaluates the experience-driven evolution capability of the proposed CDT architecture under different task scales. To this end, we measure the average search overhead required to identify the first feasible solution, which reflects how accumulated operational experience influences reasoning efficiency and task processing. A larger task scale corresponds to a more diverse task pool with different task types, priorities, deadlines, and state-dependent constraints, rather than repeated identical requests. The search overhead is defined as the number of candidate solutions examined in sequence before a feasible instruction is identified. All configurations share the same initial task-related knowledge base, and the maximum search overhead is capped at 10; tasks without a feasible solution within this limit are recorded as 10. The results show that DT-S, which relies on a static knowledge base, maintains a relatively stable search overhead across task scales without a noticeable reduction. CDT-NA shows comparable performance under small task scales, but its overhead gradually decreases as task experience accumulates through experience-driven cognitive updating. However, without access to adjacent shared knowledge and the associated inference support, its overhead remains higher than that of CDT (Case 1) and CDT (Case 2). These results show that the proposed CDT can progressively reduce search overhead through accumulated experience, confirming its ability to improve task-processing efficiency over time.

\section{Conclusions and Future Research Directions}
In this paper, we move toward an operational definition of Cognitive Digital Twins (CDTs) by proposing a four-layer self-evolving closed-loop architecture consisting of the physical layer, digital-twin layer, cognitive layer, and task layer. The proposed architecture organizes physical-state generation, digital representation, cognitive-model construction, and task-oriented decision making as distinct operational functions to enable closed-loop operation under constrained semantic information and dynamic conditions. We further describe the core workflow and key mechanisms, emphasizing experience- and knowledge-guided cognition, feasibility-aware decision generation, and feedback-driven cognitive refinement. A lightweight multi-task simulation study illustrates improved closed-loop feasibility and reduced search overhead as task experience accumulates. While these results support the effectiveness of the proposed architecture, several important open issues still require further investigation.
\begin{itemize}
    \item[1)] \textbf{Uncertainty modeling and safety boundaries}: Wireless communication and sensory inputs are inherently uncertain. Incorporating uncertainty into feasible-solution construction and analyzing the resulting closed-loop attack surface remain important directions.
    \item[2)] \textbf{Scalable large-scale implementation}: Although the architecture is illustrated at a limited scale, it faces scalability challenges in distributed knowledge graphs, cross-edge consistency, and orchestration overhead, motivating research on partitioned KGs, edge caching, and cross-layer offloading.
\end{itemize}

\bibliographystyle{ieeetr}
\bibliography{reference}

\end{document}